# Fuzzy Gesture Expression Model for an Interactive and Safe Robot Partner

Alexis Stoven-Dubois

Polytech Paris-UPMC
Pierre et Marie Curie University
4 place Jussieu, Paris, 75012, France
alexis.stoven-dubois@etu.upmc.fr

Janos Botzheim

Graduate School of System Design
Tokyo Metropolitan University
6-6 Asahigaoka, Hino, Tokyo, 191-0065 Japan
dr.janos.botzheim@ieee.org

Naoyuki Kubota

Graduate School of System Design
Tokyo Metropolitan University
6-6 Asahigaoka, Hino, Tokyo, 191-0065 Japan
kubota@tmu.ac.jp

Received October 2016; revised December 2016

ABSTRACT. Interaction with a robot partner requires many elements, including not only speech but also embodiment. Thus, gestural and facial expressions are important for communication. Furthermore, understanding human movements is essential for safe and natural interchange. This paper proposes an interactive fuzzy emotional model for the robot partner's gesture expression, following its facial emotional model.

First, we describe the physical interaction between the user and its robot partner. Next, we propose a kinematic model for the robot partner based on the Denavit-Hartenberg convention and solve the inverse kinematic transformation through Bacterial Memetic Algorithm. Then, the emotional model along its interactivity with the user is discussed. Finally, we show experimental results of the proposed model.

1. **Introduction.** Recently, the rate of elderly people is increasing in the population. Elderly people need care giving and interactive communication to be maintained in a good health state. Since the number of caregivers is not enough, other solutions have to be found. Introduction of human-friendly robot partners in the community is one possible solution.
Community-centric robots enable communication with humans. The interaction can be verbal, non-verbal, including gestures and facial expressions as well, which are very important for natural communication. For the elderly people to be entertained positively by the communication with the robots, the interaction needs to be natural [1]. As elderly people are also more fragile people getting sooner tired, it is also needed that the robot gives a secure feeling and can become less active according to the user's will.

This paper proposes a fuzzy gesture expression model based on the robot's emotional state which can change depending on the user's touching of the robot's hands. The robot is also safe as it stops every movement if encountering obstacles.
The relationship between communication and emotion has been studied in the literature [2][3] and the robot partner's emotional based communication has also been investigated and explained in our previous paper [4]. In [5], the robot's emotional state itself is entertained through communication with the user. We have also proposed a fuzzy emotional model for the face expressions based on the robot's emotional state [6]. In this paper, we propose a new gesture expression model to extend the robot's movements into a safe, fuzzy, and interactive mode, which uses the robot's emotional state as an input. For the behavior to seem natural, the gesture expression is realized according to the Laban theory of movement [7][8]. In order to apply this theory and realize the gesture expression of the robot partner, we need the kinematic and inverse kinematic model of the robot. We propose a solution for the inverse kinematic model of the robot partner using Bacterial Memetic Algorithm (BMA) [9].

This paper is organized as follows. Section 2 explains the security and interaction algorithm which enables the robot to stop its movements if encountering obstacles. Section 3 presents the kinematic model of the robot based on the Denavit-Hartenberg convention and the solution of the inverse kinematics through Bacterial Memetic Algorithm. Section 4 proposes the fuzzy gesture expression based on the robot's emotional state, which can be changed through physical interaction with the robot. Experiments are presented in Section 5. Section 6 draws the conclusions of the paper.

2. **A Safe Robot Partner.** In this paper we apply iPhonoid robot, which is a low-cost, smartphone based robot partner developed in our laboratory [4][10][16]. In order to give a secure feeling for the user, the robot needs to be able to stop its movements if it encounters an obstacle, for not breaking objects or hurting the user. As this mode is given the role of security, it needs to be always active and given priority before any other mode.
In the robot, communication is made between the smart phone and the motors through Bluetooth communication with an Arduino card. Because the Bluetooth communication may create problems due to communication speed, it was decided to include this mode directly on the Arduino card. In this way, the security feature is always directly included into the body's movements, not depending on the orders of the robot's head, which is the smartphone [10].

The movement process starts when the smartphone sends the final positions for the motors along with their speed. The Arduino card will send the order to each motor to move until their final destination with the speed demanded. During the movement process, we calculate each motor's speed by looking at their positions at periodic times. We then compare it with the speed demanded by the smart phone and look if it is under a decided limit. If the speed of only one motor is under the decided limit, the robot is considered to have encountered an obstacle and directly stops its movement. The flowchart of the blocking feature of the robot is shown in Fig. 1.

To optimize this security mode, two parameters have to be taken into account: the period $p$ and the speed limit $v$. Parameter $p$ separates the times at which we look at the motors' speed, having directly influence on the mode's reactivity but being balanced by the Arduino card's computing power. The speed limit $v$, under which we consider the robot has encountered an obstacle, needs to be high enough to give this mode enough precision, but cannot be too high in order not to detect any "false" obstacle.

As the price is one of the robot's main arguments for being inserted successfully in the society, it is equipped with cheap motors, having quite low precision concerning their position and speed. For that reason, the optimization of the security mode could only be made through real-situation tests, trying many different limit functions and different periods. By looking at the robot's geometry, it is very unlikely for it to be blocked somewhere else than at the upper arms. As a consequence, using a simple speed limit could give a sufficiently reactive and precise behavior.

The best operation has been established with the following values:

$$p = 0.1 \; second \tag{1}$$

$$v = 0.9372 \cdot v_{sent} \tag{2}$$

where $v_{sent}$ is the speed sent to the respective motor.

Finally, this mode enables physical interaction with the robot. Due to the leverage, the effort needed to stop the robot's movement by applying a force on the top of its arms is very low: putting our hand on the robot's way without caring is sufficient. As a matter of fact, it is sufficient for elderly people to physically interact with this robot without hurting themselves.

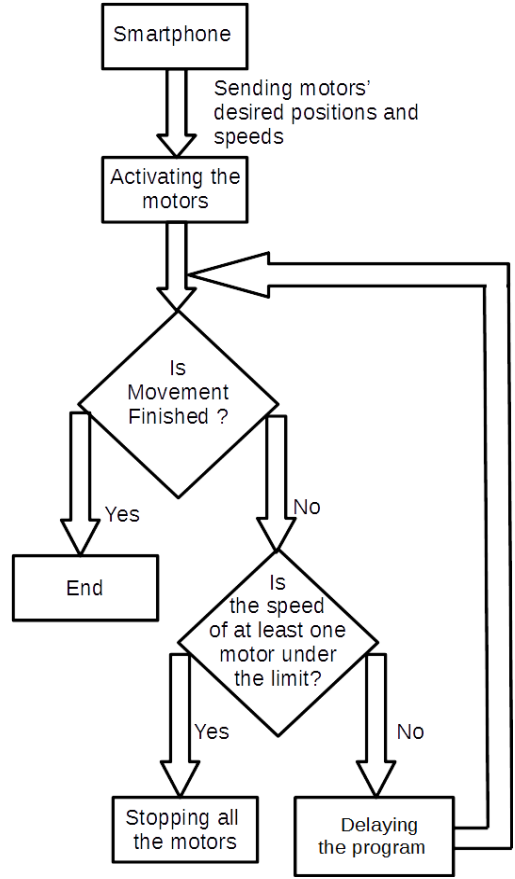

FIGURE 1. Flowchart of the blocking feature of the robot

3. **Kinematic model.** The kinematic model of the robot illustrated in Fig. 2 has been realized according to the Denavit-Hartenberg convention [11]. The static lengths of the robot are $L_{01}$, $L_{12}$, $L_{23}$, $L_{34}$, $L_{H1}$ and $L_{H2}$.

The robot has 8 degrees of freedom (DoF), and the eight motors' rotation are represented by eight angles, $\theta_W$, $\theta_{RS}$, $\theta_{RE}$, $\theta_{RF}$, $\theta_{LS}$, $\theta_{LE}$, $\theta_{LF}$, $\theta_H$ corresponding respectively to the rotations of the waist, the right shoulder, the right elbow, the right fingertip, the left shoulder, the left elbow, the left fingertip, and the head. The limitations of angles are presented in Table 1 in radians.

TABLE 1. Motors range

| Parameters | Minimum Range | Maximum Range |
|---|---|---|
| $\theta_W$ | −1.744 | 1.744 |
| $\theta_{RS}$ | −0.523 | 3.1415 |
| $\theta_{RE}$ | −0.174 | 1.744 |
| $\theta_{RF}$ | −0.174 | 1.744 |
| $\theta_{LS}$ | −0.523 | 3.1415 |
| $\theta_{LE}$ | −1.744 | 0.174 |
| $\theta_{LF}$ | −1.744 | 0.174 |
| $\theta_H$ | 0.174 | 1.22 |

The limitations have been decided so that the robot can only make movements similar to those the human articulations can realize. In that way, independently from its feelings, the robot's movements will seem natural to the user.

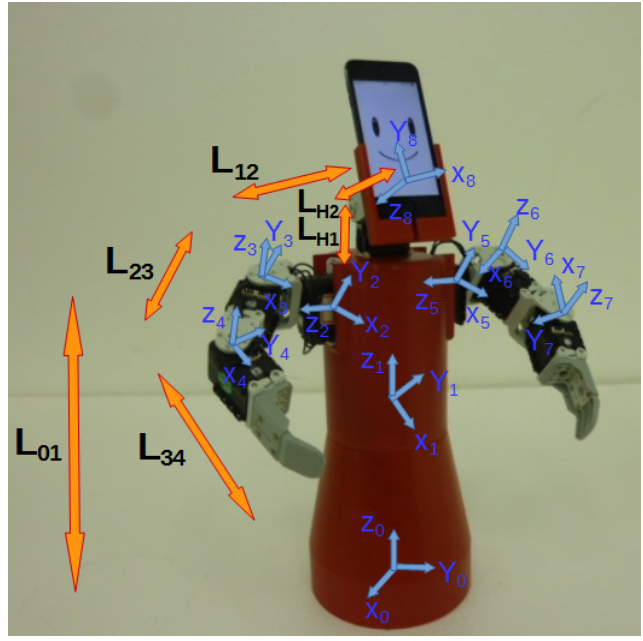

Figure 2. The robot's kinematic model

For the direct kinematic model [12], we have considered the eight angles as the inputs. The outputs are the 9 coordinates representing the positions of the two fingertips and the head center, calculated in the base coordinate system $\mathcal{R}_0$. The kinematic model is presented in Equations (3)–(11), where $c_i = cos(i)$ and $s_i = sin(i)$.

$$x_{righthand} = L_{12} \cdot s_{\theta_W} + L_{23} \cdot (c_{\theta_{RE}} \cdot s_{\theta_W} + c_{\theta_W} \cdot s_{\theta_{RS}} \cdot s_{\theta_{RE}}) \\ + L_{34} \cdot (-c_{\theta_{RF}} \cdot (-c_{\theta_{RE}} \cdot s_{\theta_W} + c_{\theta_W} \cdot s_{\theta_{RS}} \cdot s_{\theta_{RE}}) + (c_{\theta_W} \cdot c_{\theta_{RE}} \cdot s_{\theta_{RS}} - s_{\theta_W} \cdot s_{\theta_{RE}}) \cdot s_{\theta_{RF}}) \tag{3}$$

$$y_{righthand} = -L_{12} \cdot c_{\theta_W} + L_{23} \cdot (-c_{\theta_{RE}} \cdot c_{\theta_W} + s_{\theta_W} \cdot s_{\theta_{RS}} \cdot s_{\theta_{RE}}) \\ + L_{34} \cdot (-c_{\theta_{RF}} \cdot (c_{\theta_{RE}} \cdot c_{\theta_W} + s_{\theta_W} s_{\theta_{RS}} \cdot s_{\theta_{RE}}) + (s_{\theta_W} \cdot c_{\theta_{RE}} \cdot c_{\theta_W} + c_{\theta_W} \cdot s_{\theta_{RE}}) \cdot s_{\theta_{RF}}) \tag{4}$$

$$z_{righthand} = L_{01} - L_{23} \cdot c_{\theta_{RS}} \cdot s_{\theta_{RE}} + L_{34} \cdot (-c_{\theta_{RS}} \cdot c_{\theta_{RF}} \cdot s_{\theta_{RE}} - c_{\theta_{RS}} \cdot c_{\theta_{RE}} \cdot s_{\theta_{RF}}) \tag{5}$$

$$x_{lefthand} = -L_{12} \cdot s_{\theta_W} + L_{23} \cdot (-c_{\theta_{LE}} \cdot s_{\theta_W} - c_{\theta_W} \cdot s_{\theta_{LS}} \cdot s_{\theta_{LE}}) \\ + L_{34} \cdot (c_{\theta_{LF}} \cdot (-c_{\theta_{LE}} \cdot s_{\theta_W} + c_{\theta_W} \cdot s_{\theta_{LS}} \cdot s_{\theta_{LE}}) - (c_{\theta_W} \cdot c_{\theta_{LE}} \cdot s_{\theta_{LS}} - s_{\theta_W} \cdot s_{\theta_{LE}}) \cdot s_{\theta_{LF}}) \tag{6}$$

$$y_{lefthand} = L_{12} \cdot c_{\theta_W} + L_{23} \cdot (c_{\theta_{LE}} \cdot c_{\theta_W} - s_{\theta_W} \cdot s_{\theta_{LS}} \cdot s_{\theta_{LE}}) \\ + L_{34} \cdot (c_{\theta_{LF}} \cdot (c_{\theta_{LE}} \cdot c_{\theta_W} + s_{\theta_W} s_{\theta_{LS}} \cdot s_{\theta_{LE}}) - (s_{\theta_W} \cdot c_{\theta_{LE}} \cdot c_{\theta_W} + c_{\theta_W} \cdot s_{\theta_{LE}}) \cdot s_{\theta_{LF}}) \tag{7}$$

$$z_{lefthand} = -L_{01} + L_{23} \cdot c_{\theta_{LS}} \cdot s_{\theta_{LE}} + L_{34} \cdot (c_{\theta_{LS}} \cdot c_{\theta_{LF}} \cdot s_{\theta_{LE}} + c_{\theta_{LS}} \cdot c_{\theta_{LE}} \cdot s_{\theta_{LF}}) \tag{8}$$

$$x_{headcenter} = L_{H2} \cdot c_{\theta_W} \cdot c_{\theta_H} \tag{9}$$

$$y_{headcenter} = L_{H2} \cdot s_{\theta_W} \cdot c_{\theta_H} \tag{10}$$

$$z_{headcenter} = L_{01} + L_{H1} + L_{H2} \cdot c_{\theta_H} \tag{11}$$

The Iphonoid's kinematic representation is shown in Fig. 3. The cylinders represent pivot links. $S_1$, $S_2$, $S_3$, $S_4$, $S_5$, $S_6$, $S_7$, $S_8$ correspond respectively to the waist, right shoulder, right elbow, right fingertip, left shoulder, left elbow, left fingertip and head of the robot.
In order to reach some desired positions by the robot and describe the robot's gestures by angles, the inverse kinematics of the robot has to be solved. This would enable us to get the 8 angles of the motors as outputs, having the 9 coordinates representing the positions of the robot's right arm, left arm, and head in the 3D space as inputs. In this way, we could control the robot by the position of its arms and head described in the Cartesian space. However, the robot possesses 2 parallel links in each arm: the elbow and the fingertip, as shown in Fig. 3. As a consequence, its arms theoretically can go to a same position by different ways. The mathematical solution becomes indeed more complicated since the direct inversion of the forward kinematic model gives us 8 solutions for the inverse kinematic model. However, considering all the limitations we have artificially put onto each articulation, there should be only one solution for getting each extremity to a desired and accessible position. In this paper, we propose a novel solution for the inverse kinematic problem. In order to solve the inverse kinematic problem in real time, we describe it as an optimization problem and apply an evolutionary approach, the Bacterial Memetic Algorithm to solve it.

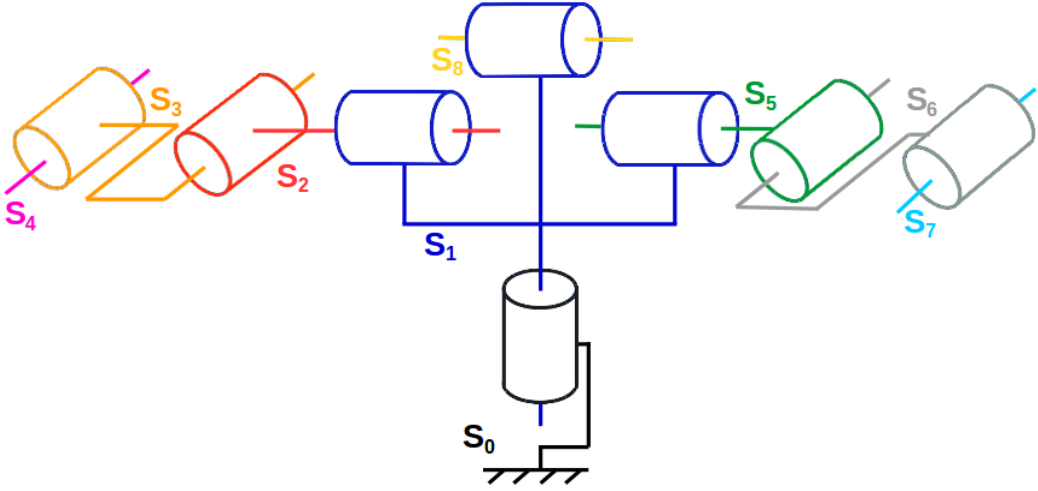

FIGURE 3. The robot's kinematic representation

Bacterial Memetic Algorithm [9] is a population based stochastic optimization technique which effectively combines global and local search in order to find good quasi-optimal solution for the given problem. In the global search, BMA applies the bacterial operators, the bacterial mutation and the gene transfer operation. The role of the bacterial mutation is the optimization of the bacteria's chromosome. The gene transfer allows the information's transfer in the population among the different bacteria. As a local search technique, we apply the Levenberg-Marquardt method [13][14] by a certain probability for each individual [9].

The operation of the BMA starts with the generation of a random initial population containing $N_{ind}$ individuals. Next, until a stopping criterion is fulfilled (which is usually the number of generations, $N_{gen}$), we apply the bacterial mutation, local search, and gene transfer operators.

Bacterial mutation creates $N_{clones}$ number of clones (copies) of an individual, which are then subjected to random changes in their genes. The number of genes that are modified with this mutation is a parameter of the algorithm ($l_{bm}$). This mutation step is performed on all clones of the individual except one, which is kept the same as the original. After mutating the same segment in the clones, each clone is evaluated. The clone with the best evaluation result transfers the mutated segment to the other clones. These steps are repeated until each segment of the chromosome has been mutated once. This process is performed for each member of the population.

After the bacterial mutation, for each individual the Levenberg-Marquardt algorithm is used, which is a gradient based optimization technique [13][14]. The method is applied to all bacteria one by one by a given probability, $LM_{prob}$. Let $\mathbf{b}_k$ be the variables encoded in bacterium $b$ at iteration $k$. The task is to find the minimum place of the function. For a given bacterium at the $k$-th iteration the update vector is:

$$\mathbf{s}_k = -\left(\hat{J}(\mathbf{b}_k) \circ \hat{J}^T(\mathbf{b}_k) + \gamma_k \hat{I}\right)^{-1} \hat{J}(\mathbf{b}_k), \qquad (12)$$

where $\hat{J}(\mathbf{b}_k)$ is the gradient vector of $f$ evaluated at $\mathbf{b}_k$, $\gamma_k$ is a parameter initially set to any positive value ($\gamma_{init} = \gamma_1 > 0$), $\hat{I}$ is the identity matrix and $\circ$ stands for the dyadic product of two vectors. The derivatives in $\hat{J}(\mathbf{b}_k)$ are approximated by second-order accurate, central finite differences. After the update vector was computed we calculate the so-called trust region, $r_k$ as follows: $r_k = \frac{f(\mathbf{b}_k+\mathbf{s}_k)-f(\mathbf{b}_k)}{\hat{J}(\mathbf{b}_k)^T \cdot \mathbf{s}_k}$. The value of parameter $\gamma$ is adjusted dynamically depending on the value of $r_k$:

- If $r_k < 0.25$ then $\gamma_{k+1} = 4\gamma_k$
- If $r_k > 0.75$ then $\gamma_{k+1} = \gamma_k/2$
- Else $\gamma_{k+1} = \gamma_k$

If $f(\mathbf{b}_k + \mathbf{s}_k) < f(\mathbf{b}_k)$ then $\mathbf{b}_{k+1} = \mathbf{b}_k + \mathbf{s}_k$, else $\mathbf{b}_{k+1} = \mathbf{b}_k$. If the stopping condition ($||\hat{J}(\mathbf{b}_k)|| \leq \tau$) is fulfilled or a predefined maximum iteration number ($LM_{iter}$) is reached then the Levenberg-Marquardt algorithm stops, otherwise it continues with the $(k+1)$-th iteration step.

The last operator in a generation is the horizontal gene transfer. It means copying genes from better individuals to worse ones. For this reason, the population is split into two halves, according to the cost values. One randomly chosen bacterium from the better half of the population transfers $l_{gt}$ genes to another randomly chosen bacterium from the worst half of the population. After transferring the genes, the population is sorted again. The number of gene transfers in one generation is $N_{inf}$.

Bacterial memetic algorithm has been successfully applied to a wide range of problems. More details about the algorithm can be found e.g. in [9][15].

In the case of evolutionary and memetic algorithms we have to discuss the encoding method and the evaluation of the individuals (bacteria) as well. In our case, each bacterium consists of 8 real numbers each one describing the corresponding motor's angle. The evaluation of the bacteria is calculated by Eq. (13).

$$\begin{aligned}
E = &(x_{righthand}^{desired} - x_{righthand}^{calculated})^2 + (y_{righthand}^{desired} - y_{righthand}^{calculated})^2 + (z_{righthand}^{desired} - z_{righthand}^{calculated})^2 \\
&+ (x_{lefthand}^{desired} - x_{lefthand}^{calculated})^2 + (y_{lefthand}^{desired} - y_{lefthand}^{calculated})^2 + (z_{lefthand}^{desired} - z_{lefthand}^{calculated})^2 \\
&+ (x_{headcenter}^{desired} - x_{headcenter}^{calculated})^2 + (y_{headcenter}^{desired} - y_{headcenter}^{calculated})^2 + (z_{headcenter}^{desired} - z_{headcenter}^{calculated})^2
\end{aligned} \quad (13)$$

where the variables with *desired* superscript mean the desired coordinate for the given extremity, and the variables having *calculated* superscript mean the calculated coordinate by the forward kinematics using the values encoded in the given bacterium as motor values.

For each extremity, we calculate the distance from its desired position the extremity would be if we chose the values encoded in the bacterium as motor values. Using the sum of these differences, $E$ represents how far from the desired solution the bacterium is. By eliminating the far bacteria and giving reproduction chance to closer bacterium, we can finally obtain an optimal solution.

The parameter setting used in the BMA is presented in Table 2. We found these parameters by making real-situation tests in order to get always a solution with 0.015 radian precision within 1 second on the smart phone.

Table 2. BMA Parameters

| Parameter | $N_{gen}$ | $N_{ind}$ | $N_{clones}$ | $l_{bm}$ | $N_{inf}$ |
|---|---|---|---|---|---|
| Value | 35 | 12 | 10 | 1 | 15 |
| Parameter | $l_{gt}$ | $LM_{prob}$ | $LM_{iter}$ | $\gamma_{init}$ | $\tau$ |
| Value | 1 | 20% | 8 | 1.0 | 0.0001 |

4. **Fuzzy Gesture Expression Model.** The robot's emotional model is described by three values: the emotion, the feeling, and the mood [4][6][16]. The emotion is a short-time state influenced by the very last events during the shortest scale of time. The feeling is a longer-time state influenced by events on a longer scale of time. The mood is the longest-time state influenced by events on the longest scale of time.

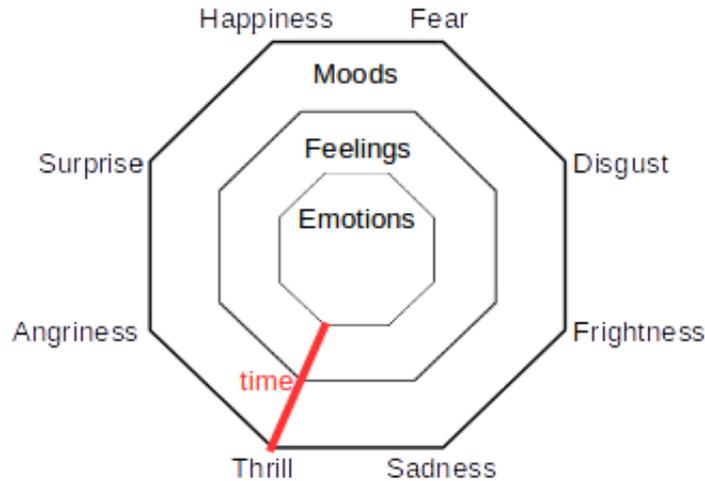

FIGURE 4. Iphonoid's emotional model

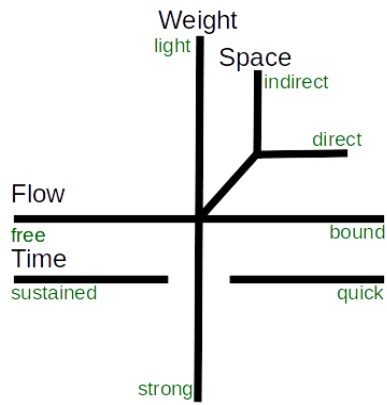

FIGURE 5. Laban effort graph

These three values will be influenced when communicating with the user [4][6][16]. When speaking to a happy user, the robot will automatically get a happy emotion, but it will take a longer time speaking with a happy user to get a happy mood.

The robot have 8 main feelings: happiness, sadness, fright, fear, thrill, disgust, angriness, and surprise. The one with the strongest intensity at the moment becomes the one the robot feels [16]. The Iphonoid's emotional model is illustrated in Fig. 4. In this paper, we propose a fuzzy gesture expression model based on the robot's feeling.

The robot is given 8 movements functions for each extremity of the robot. Each function corresponds to one of the 8 main feelings. The model takes an emotional intensity as an input and gives 9 coordinates for the positions of the robot's 3 extremities as an output. The general shapes of the functions have been designed following humans' with a similar feeling. For the robot to take into account the emotional intensity in a natural way, we used the Laban theory of movement. This theory describes every human movement with four parameters, as illustrated in Fig. 5. Following this theory, the more a person's feelings are intense, the more his/her movements will use space and will be fast [7][8]. Thus, the direction and speed of each movement function have been designed using the space and the time parameter.

First, we take the robot's emotion value to decide which movement function to choose. As the emotion is the shortest emotional state of the robot, it is also the most intense, and so, the state it will clearly express to the user [17]. Next, we take the robot's mood value as the emotional intensity for the movement function chosen beforehand. The mood's value is the longest emotional state of the robot. It is created by a repetition of the same emotions on a longer scale of time. In this way, the mood represents a less intense emotional state but having more effect on the robot's state of mind on a longer scale of time. For that reason, it is natural for it not to have any influence on the robot's current type of gesture but to have a great influence on the emotional intensity shown by the robot. The flowchart of the fuzzy gesture expression model is described in Fig. 6.

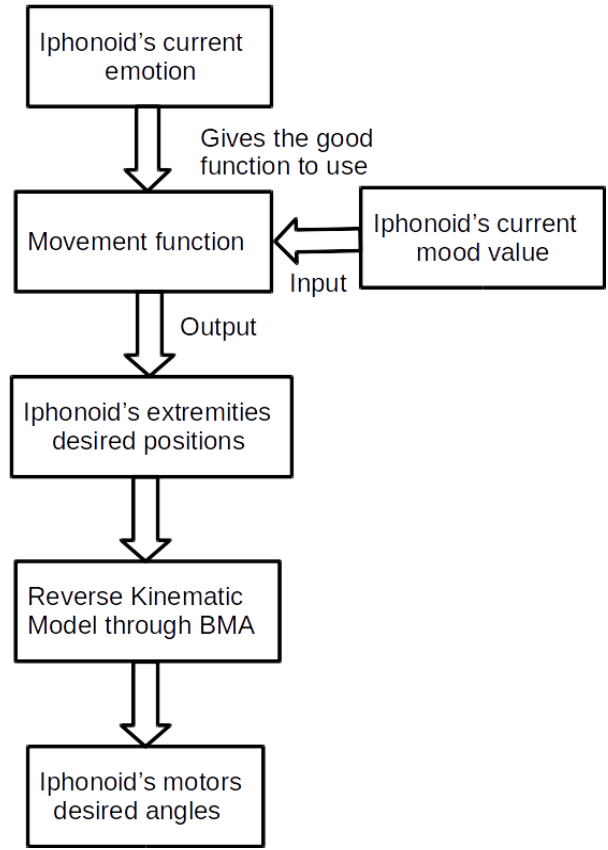

FIGURE 6. Flowchart of the fuzzy gesture model

Then, the movement function corresponding to the current emotion is used with the mood value $\mu$ as an input and it gives the 9 coordinates for the robot's extremities as an output. The mood value $\mu$ is considered as a fuzzy membership value for the given feeling as depicted in Fig. 7. The proposed movement functions are illustrated in Fig. 7.

Finally, the inverse kinematic model through the Bacterial Memetic Algorithm is used to move the robot in accordance with the Laban Theory. Indeed, each movement function is defined so that the greater the mood value is, the more space the movement will use, and the faster it will be executed. This way, each movement will seem natural and in accordance with the robot's feelings to the user.

Our gesture model has also been given the possibility of physical interaction using the blocking-security mode described in Section 2. Whenever the robot touches an obstacle, it stops all of its movements. However, it is also natural for the robot, whatever type of contact it is, object or human, witting or unwitting, to have its behavior changing after the contact.

The choice that has been made for our gesture expression model is to decrease the mood value by 20% after every encounter. The decrease will then be able to be erased within some time, at the same rhythm as the mood value. Thus, the robot will reduce its feeling's intensity following the Laban theory when it touches an obstacle and then it will come back into the neutral position. If the contact was unwitting, the robot will reduce its emotional state because it will consider itself to have taken too much risks and to have made a mistake by touching its environment. If the contact was established by the user's will to block the robot's movements, the robot will understand it as a request from the user to lower its activity, and so it will lower its emotional state. As the robot is destined for elders who get easily injured or tired, it is natural for the robot to consider this two types of information as important information, and so, to have them influence directly its most difficult value to influence: the mood.

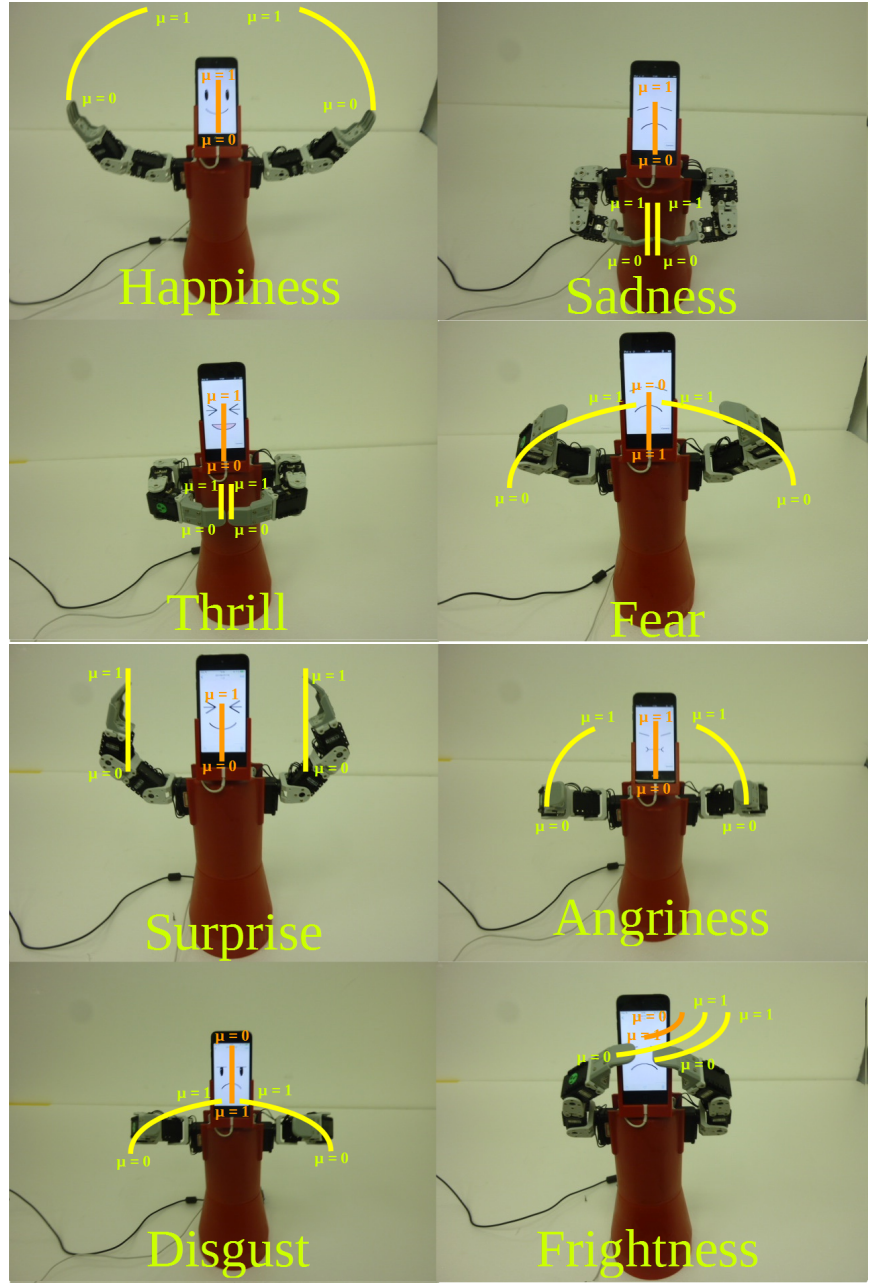

Figure 7. Robot's movement functions

5. **Experiments.** In the experiment a happy emotion with a mood value of 0.9 has been set at the start. Then the robot periodically moves with a period of $p = 2s$, depending on its emotion and mood value. At some point, the user is blocking the robot's movements in order to decrease the mood value.

The gesture expression model and the secure-blocking worked as expected. The inverse kinematic calculation by the Bacterial Memetic Algorithm was working within 1 second of calculation and the robot moved to the right positions according to the happiness movement function and its mood value.

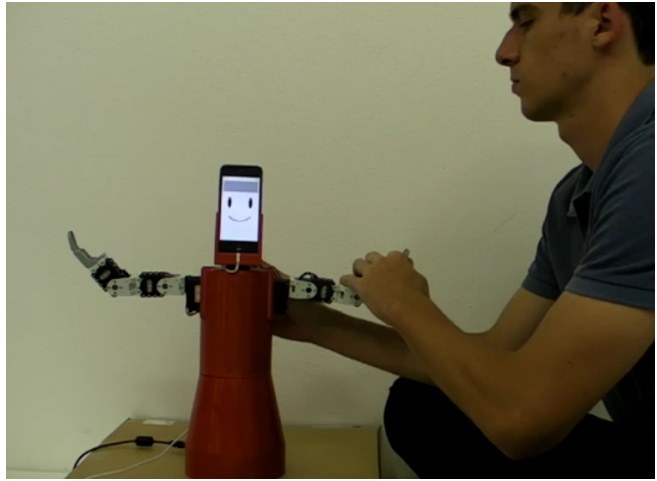

Figure 8. Fuzzy gesture expression model experiment

The robot was also stopping its movement, coming back to neutral position, and decreasing its mood value whenever and only when its movements were blocked. As a consequence, its movements were then slower and used less space, according to the Laban Theory. The experiment is illustrated in Fig. 8 and its video is available at http://alexis.stoven-dubois.delfosia.net/Iphonoid.

6. **Conclusion.** This paper proposed a fuzzy gesture expression model depending on feeling values directly influenced through communication with the user. The robot possesses 8 main movement functions for every of its main feelings, and by using an inverse kinematic model through the Bacterial Memetic Algorithm, moves accordingly to its current emotion and mood. As every movement function follows the Laban theory, and as the robot's articulations can only produce human-like rotations, the robot's movements always seem natural and in accordance with its current emotional state. It also has a secure behavior and enables easy physical interaction to lower its activity. Adding this to its cheap price by using a smart phone as the robot's head makes it accessible and useful for elder people to keep daily and natural communication.